\documentclass[10pt,twocolumn,letterpaper]{article}

\usepackage{cvpr}
\usepackage{times}
\usepackage{epsfig}
\usepackage{graphicx}
\usepackage{amsmath}
\usepackage{amssymb}
\usepackage{algorithm}
\usepackage{algorithmicx}
\usepackage[noend]{algpseudocode}
\usepackage{color}

\usepackage[breaklinks=true,bookmarks=false]{hyperref}

\cvprfinalcopy 

\def\cvprPaperID{****} 

\begin{document}

\title{Multi-Oriented Scene Text Detection via \\ Corner Localization and Region Segmentation}

\author{Pengyuan Lyu\textsuperscript{1}, Cong Yao\textsuperscript{2}, Wenhao Wu\textsuperscript{2}, Shuicheng Yan\textsuperscript{3}, Xiang Bai\textsuperscript{1}\\
\textsuperscript{1}Huazhong University of Science and Technology\\
\textsuperscript{2}Megvii Technology Inc.\\
\textsuperscript{3}National University of Singapore\\
{\tt\small \{lvpyuan, yaocong2010\}@gmail.com, wwh@megvii.com, eleyans@nus.edu.sg, xbai@hust.edu.cn}}

\maketitle

\begin{abstract}
Previous deep learning based state-of-the-art scene text detection methods can be roughly classified into two categories. The first category treats scene text as a type of general objects and follows general object detection paradigm to localize scene text by regressing the text box locations, but troubled by the arbitrary-orientation and large aspect ratios of scene text. The second one segments text regions directly, but mostly needs complex post processing. In this paper, we present a method that combines the ideas of the two types of methods while avoiding their shortcomings. We propose to detect scene text by localizing corner points of text bounding boxes and segmenting text regions in relative positions. In inference stage, candidate boxes are generated by sampling and grouping corner points, which are further scored by segmentation maps and suppressed by NMS. Compared with previous methods, our method can handle long oriented text naturally and doesn't need complex post processing. The experiments on ICDAR2013, ICDAR2015, MSRA-TD500, MLT and COCO-Text demonstrate that the proposed algorithm achieves better or comparable results in both accuracy and efficiency. Based on VGG16, it achieves an F-measure of \textbf{$84.3\%$} on ICDAR2015 and \textbf{$81.5\%$} on MSRA-TD500.
\end{abstract}

\section{Introduction}

Recently, extracting textual information from natural scene images has become increasingly popular, due to the growing   demands of real-world applications (e.g., product search \cite{bai2017integrating}, image retrieval \cite{jaderberg2016reading}, and autonomous driving). Scene text detection, which aims at locating text in natural images, plays an important role in various text reading systems \cite{neumann2010method, epshtein2010detecting, wang2011end, bissacco2013photoocr, jaderberg2014deep,gomez2017textproposals, Busta_2017_ICCV, Li_2017_ICCV}.

\begin{figure}
\vspace{-2mm}

\begin{centering}
\includegraphics[scale=0.32]{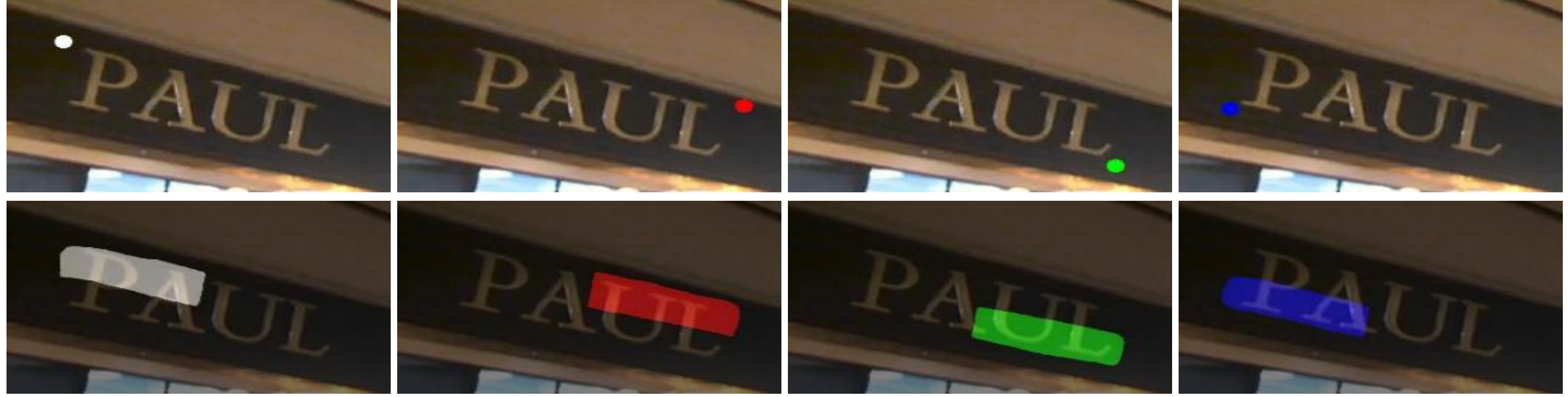}
\par\end{centering}
\caption{The images in top row and bottom row are the predicted corner points and position-sensitive maps in top-left, top-right, bottom-right, bottom-left order, respectively.}
\label{img_overview}
\vspace{-5mm}
\end{figure}

Scene text detection is challenging due to both external and internal factors. The external factors come from the environment, such as noise, blur and occlusion, which are also major problems disturbing general object detection. The internal factors are caused by properties and variations of scene text. Compared with general object detection, scene text detection is more complicated because: 1) Scene text may exist in natural images with arbitrary orientation, so the bounding boxes can also be rotated rectangles or quadrangles; 2) The aspect ratios of bounding boxes of scene text vary significantly; 3) Since scene text can be in the form of characters, words, or text lines, algorithms might be confused when locating the boundaries. 





\begin{figure*}
\vspace{-2mm}
\begin{centering}
\includegraphics[scale=0.25]{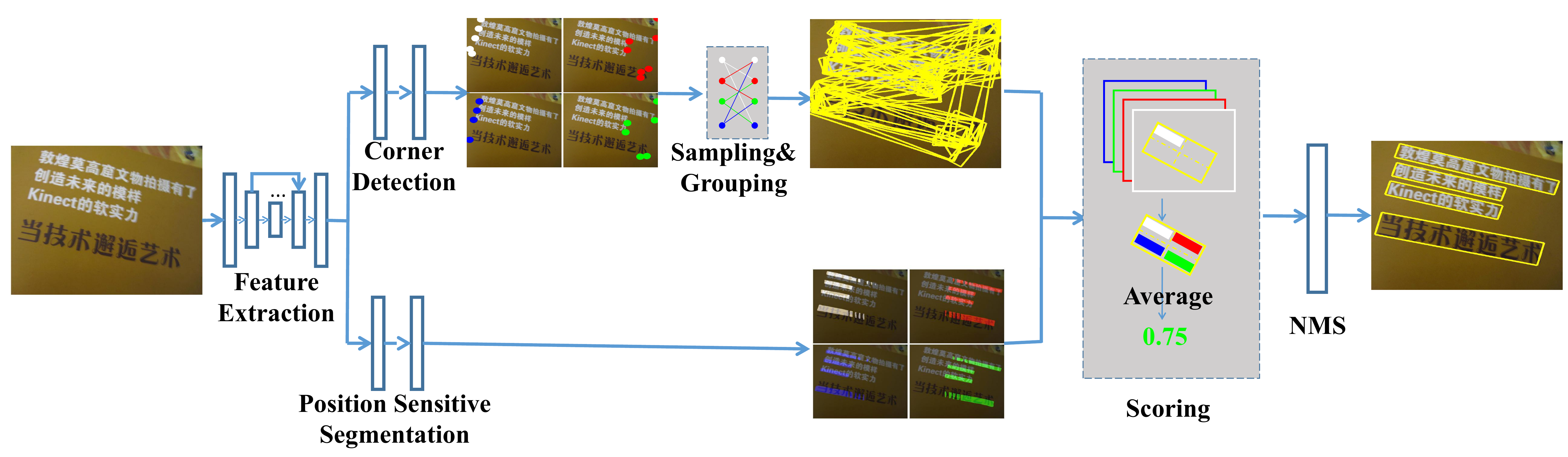}
\par\end{centering}
\caption{Overview of our method. Given an image, the network outputs corner points and segmentation maps by corner detection and position-sensitive segmentation. Then candidate boxes are generated by sampling and grouping corner points. Finally, those candidate boxes are scored by segmentation maps and suppressed by NMS.}
\label{img_pipeline}
\end{figure*}
In the past few years, scene text detection has been widely studied \cite{epshtein2010detecting, bissacco2013photoocr,yao2012detecting,jaderberg2014deep,tian2015text,zhang2016multi,Shi_2017_CVPR,tian2017wetext} and has achieved obvious progresses recently, with the rapid development of general object detection and semantic segmentation. Based on general object detection and semantic segmentation models, several well-designed modifications are made to detect text more accurately. Those scene text detectors can be split into two branches. The first branch is based on general object detectors (SSD \cite{liu2016ssd}, YOLO \cite{redmon2016you} and DenseBox \cite{huang2015densebox}), such as TextBoxes \cite{liao2017textboxes}, FCRN \cite{gupta2016synthetic}  and EAST \cite{Zhou_2017_CVPR} \emph{etc.}, which predict candidate bounding boxes directly. The second branch is based on semantic segmentation, such as \cite{zhang2016multi} and \cite{yao2016scene}, which generate segmentation maps and produce the final text bounding boxes by post-processing.

Different from previous methods, in this paper we combine the ideas of object detection and semantic segmentation and apply them in an alternative way. Our motivations mainly come from two observations: 1) a rectangle can be determined by corner points, regardless of the size, aspect ratio or orientation of the rectangle; 2) region segmentation maps can provide effective location information of text. Thus, we first detect the corner points (top-left, top-right, bottom-right, bottom-left, as shown in Fig. \ref{img_overview}) of text region rather than text boxes directly. Besides, we predict position-sensitive segmentation maps (shown in Fig. \ref{img_overview}) instead of a text/non-text map as in \cite{zhang2016multi} and \cite{yao2016scene}. Finally, we generate candidate bounding boxes by sampling and grouping the detected corner points and then eliminate unreasonable boxes by segmentation information. The pipeline of our proposed method is depicted in Fig. \ref{img_pipeline}.

The key advantages of the proposed method are as follows: 1) Since we detect scene text by sampling and grouping corner points, our approach can naturally handle arbitrary-oriented text; 2) As we detect corner points rather than text bounding boxes, our method can spontaneously avoid the problem of large variation in aspect ratio; 3) With position-sensitive segmentation, it can segment text instances well, no matter the instances are characters, words, or text lines; 4) In our method, the boundaries of candidate boxes are determined by corner points. Compared with regressing text bounding box from anchors ( \cite{liao2017textboxes, ma2017arbitrary}) or from text regions (\cite{Zhou_2017_CVPR,He_2017_ICCV}), the yielded bounding boxes are more accurate, particularly for long text.

We validate the effectiveness of our method on horizontal, oriented, long and oriented text as well as multi-lingual text from public benchmarks. The results show the advantages of the proposed algorithm in accuracy and speed. Specifically, the F-Measures of our method on ICDAR2015 \cite{karatzas2015icdar}, MSRA-TD500 \cite{yao2012detecting} and MLT \cite{MLT-Challenge} are \textbf{$84.3\%$}, \textbf{$81.5\%$} and \textbf{$72.4\%$} respectively, which outperform previous state-of-the-art methods significantly. Besides, our method is also competitive in efficiency. It can process more than \textbf{10.4} images (512x512 in size) per second. 

The contributions of this paper are four-fold: (1) We propose a new scene text detector that combines the ideas of object detection and segmentation, which can be trained and evaluated end-to-end. (2) Based on position-sensitive ROI pooling \cite{dai2016r}, we propose a rotated position-sensitive ROI average pooling layer that can handle arbitrary-oriented proposals. (3) Our method can simultaneously handle the challenges (such as rotation, varying aspect ratios, very close instances) in multi-oriented scene text, which are suffered by previous methods. (4) Our method achieves better or competitive results in both accuracy and efficiency.

\section{Related Work}

\subsection{Regression Based Text Detection}

Regression based text detection has become the mainstream of scene text detection in the past two years. Based on general object detectors, several text detection methods were proposed and achieved substantial progress. Originating from SSD \cite{liu2016ssd}, TextBoxes \cite{liao2017textboxes} use "long" default boxes and "long" convolutional filters to cope with the extreme aspect ratios. Similarly, in \cite{ma2017arbitrary} Ma \emph{et al.} utilize the architecture of Faster-RCNN \cite{ren2015faster} and add rotated anchors in RPN to detect arbitrary-oriented scene text. SegLink \cite{Shi_2017_CVPR} predicts text segments and the linkage of them in a SSD style network and links the segments to text boxes, in order to handle long oriented text in natural scene. Based on DenseBox \cite{huang2015densebox}, EAST \cite{Zhou_2017_CVPR} regresses text boxes directly.

Our method is also adapted from a general object detector DSSD \cite{fu2017dssd}. But unlike the above methods that regress text boxes or segments directly, we propose to localize the positions of corner points, and then generate text boxes by sampling and grouping the detected corners.

\subsection{Segmentation Based Text Detection}

Segmentation based text detection is another direction of text detection. Inspired by FCN \cite{long2015fully}, some methods are proposed to detect scene text by using segmentation maps. In \cite{zhang2016multi}, Zhang \emph{et al.} first attempt to extract text blocks from a segmentation map by a FCN. Then they detect characters in those text blocks with MSER \cite{neumann2010method} and group the characters to words or text lines by some priori rules. In \cite{yao2016scene}, Yao \emph{et al.} use a FCN to predict three types of maps (text regions, characters, and linking orientations) of the input images. Then some post-processings are conducted to obtain text bounding boxes with the segmentation maps.

Different from the previous segmentation based text detection methods, which usually need complex post-processing, our method is simpler and clearer. In inference stage, the position-sensitive segmentation maps are used to score the candidate boxes by our proposed \textbf{Rotated Position-Sensitive Average ROI Pooling} layer.


\subsection{Corner Point Based General Object Detection}

Corner point based general object detection is a new stream of general object detection methods. In DeNet \cite{Tychsen-Smith_2017_ICCV}, Tychsen-Smith \emph{et al.} propose a corner detect layer and a sparse sample layer to replace RPN in a Faster-RCNN style two-stage model. In \cite{wang2017point}, Wang \emph{et al.} propose PLN (Point Linking Network) which regresses the corner/center points of bounding-box and their links using a fully convolutional network. Then the bounding boxes of objects are formed using the corner/center points and their links.

Our method is inspired by those corner point based object detection methods, but there are key differences. First, the corner detector of our method is different. Second, we use segmentation map to score candidate boxes. Third, it can produce arbitrary-oriented boxes for objects (text).


\subsection{Position-Sensitive Segmentation}

Recently, instance-aware semantic segmentation methods are proposed with position-sensitive maps. In \cite{dai2016instance}, Dai \emph{et al.} first introduce relative position to segmentation and propose InstanceFCN for instance segment proposal. In FCIS \cite{Li_2017_CVPR}, with the assistance of position-sensitive inside/outside score maps, Li \emph{et al.} propose an end-to-end network for instance-aware semantic segmentation.


We also adopt position-sensitive segmentation maps to predict text regions. Compared with the above-mentioned methods, there are three key differences: 1) We optimize the network with position-sensitive ground truth directly (detailed in Sec  \ref{sec_label}); 2)  Our position-sensitive maps can be used to predict text regions and score proposals simultaneously (detailed in Sec \ref{sec_scoring}), different from FCIS which uses two types of   position-sensitive maps (inside and outside); 3) Our proposed Rotated Position-Sensitive ROI Average Pooling can handle arbitrary-oriented proposals.

\begin{figure*}
\vspace{-3mm}

\begin{centering}
\includegraphics[scale=0.4]{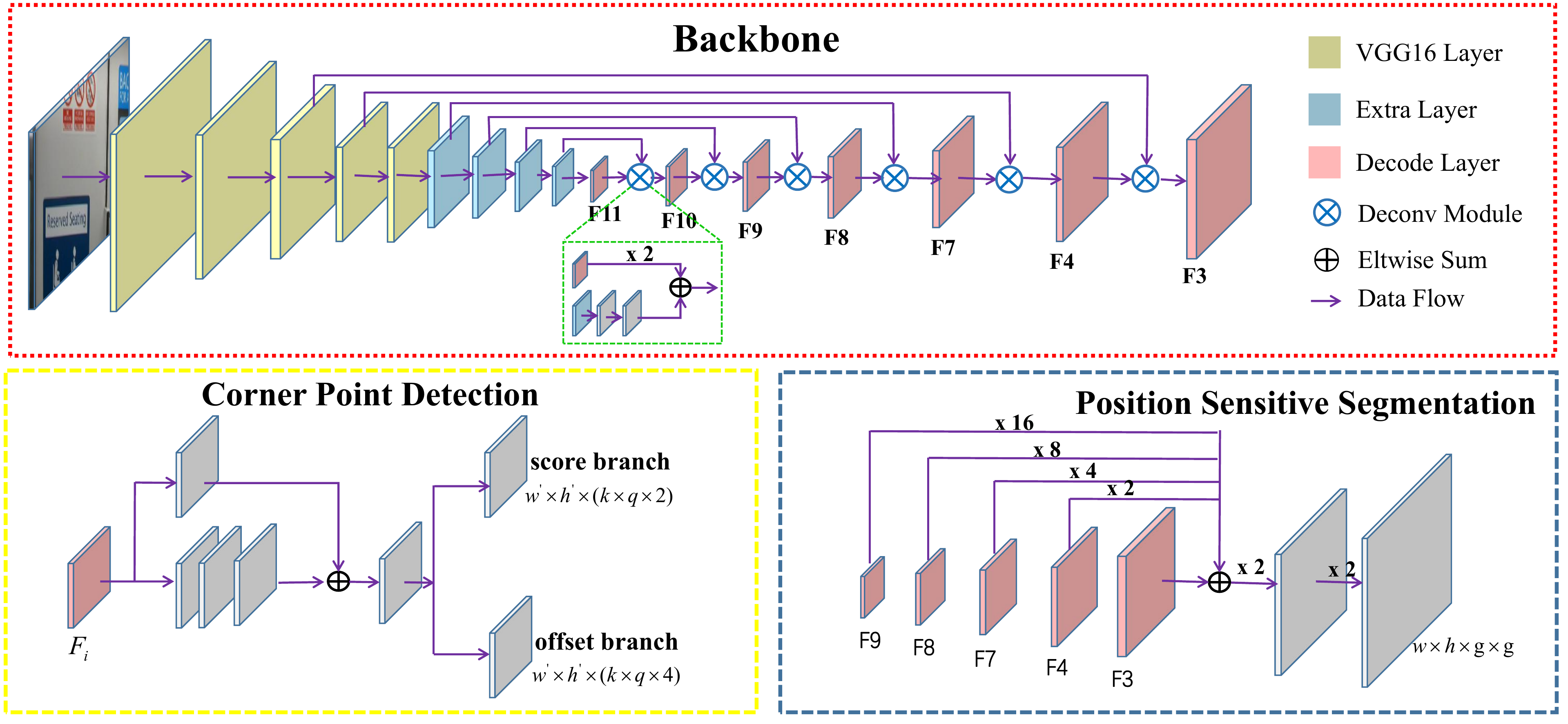}
\par\end{centering}
\caption{Network Architecture. The network contains three parts: backbone, conner point detector and position-sensitive segmentation predictor. The backbone is adapted from DSSD \cite{fu2017dssd}. Conner point detectors are built on multiple feature layers (blocks in pink). position-sensitive segmentation predictor shares some features (pink blocks) with corner point detectors.}
\label{img_network}
\end{figure*}

\section{Network}

The network of our method is a fully convolutional network that plays the roles of feature extraction, corner detection and position-sensitive segmentation. The network architecture is shown in Fig. \ref{img_network}. Given an image, the network produces candidate corner points and segmentation maps.

\subsection{Feature Extraction}

The backbone of our model is adapted from a pre-trained VGG16 \cite{simonyan2014very} network and designed with the following considerations: 1) the size of scene text varies hugely, so the backbone must has enough capacity to handle this problem well; 2) backgrounds in natural scenes are complex, so the features should better contain more context. Inspired by the good performance achieved on those problem by FPN \cite{Lin_2017_CVPR} and DSSD \cite{fu2017dssd}, we adopt the backbone in FPN/DSSD architecture to extract features. 

In detail, we convert the fc6 and fc7 in the VGG16 to convolutional layers and name them conv6 and conv7 respectively. Then several extra convolutional layers (conv8, conv9, conv10, conv11) are stacked above conv7 to  enlarge the receptive fields of extracted features. After that, a few deconvolution modules proposed in DSSD \cite{fu2017dssd} are used in a top-down pathway (Fig. \ref{img_network}). Particularly, to detect text with different sizes well, we cascade deconvolution modules with $256$ channels  from conv11 to conv3 (the features from conv10, conv9, conv8, conv7, conv4, conv3 are reused), and 6 deconvolution modules are built in total. Including the features of conv11, we name those output features $F_{3}, F_{4}, F_{7}, F_{8}, F_{9}, F_{10}$ and $F_{11}$ for convenience. In the end, the feature extracted by conv11 and deconvolution modules which have richer feature representations are used to detect corner points and predict position-sensitive maps.

\subsection{Corner Detection}

For a given rotated rectangular bounding box $R=(x,y,w,h,\theta)$, there are 4 corner points (top-left, top-right, bottom-right, bottom-left) and can be represented as two-dimensional coordinates $\{(x_{1},y_{1}),(x_{2},y_{2}),(x_{3},y_{3}),(x_{4},y_{4})\}$ in a clockwise direction. To expediently detect corner points, here we redefine and represent a corner point by a horizontal square $C=(x_{c},y_{c},ss,ss)$, where $x_{c},y_{c}$ are the coordinate of a corner point (such as $x_{1},y_{1}$ for top-left point) as well as  the center of the horizontal square. $ss$ is the length short side of the rotated rectangular bounding box $R$. 

Following SSD and DSSD, we detect corner points with default boxes. Different from the manner in SSD or DSSD where each default box outputs the classification scores and offsets of the corresponding candidate box, corner point detection is more complex because there might be more than one corner points in the same location (such as a location can be the bottom-left corner and top-right corner of two boxes simultaneously). So in our case, a default box should output classification scores and offsets for $4$ candidate boxes corresponding to the $4$ types of corner points. 

We adapt the prediction module proposed in \cite{fu2017dssd} to predict scores and offsets in two branches in a convolutional manner. In order to reduce the computational complexity, the filters of all convolutions are set to $256$. For an $m\times n$ feature map with $k$ default boxes in each cell, the "score" branch and "offset" branch output $2$ scores and $4$ offsets respectively  for each type of corner point of each default box. Here, $2$ for "score" branch means whether a corner point exists in this position. In total, the output channels of the "score" branch and the "offset" branch  are $k\times q\times2$ and $k\times q\times4$, where $q$ means the type of corner points. By default, $q$ is equal to 4. 

In the training stage, we follow the matching strategy of default boxes and ground truth ones in SSD. To detect scene text with different sizes, we use default boxes of multiple sizes on multiple layer features. The scales of all default boxes are listed in Table \ref{tab_scales}. The aspect ratios of default boxes are set to $1$. 

\begin{table*}
\vspace{-2mm}

\centering{}%
\begin{tabular}{|c|c|c|c|c|c|c|c|}
\hline 
{\small{}layer} & {\small{}$F_{3}$} & {\small{}$F_{4}$} & {\small{}$F_{7}$} & {\small{}$F_{8}$} & {\small{}$F_{9}$} & {\small{}$F_{10}$} & {\small{}$F_{11}$}\tabularnewline
\hline 
\hline 
{\small{}scales} & {\small{}$4,8,6,10,12,16$} & {\small{}$20,24,28,32$} & {\small{}$36,40,44,48$} & {\small{}$56,64,72,80$} & {\small{}$88,96,104,112$} & {\small{}$124,136,148,160$} & {\small{}$184,208,232,256$}\tabularnewline
\hline 
\end{tabular}
\caption{Scales of default boxes on different layers.}
\label{tab_scales}
\end{table*}

\subsection{Position-Sensitive Segmentation}

In the previous segmentation based text detection methods \cite{zhang2016multi, yao2016scene}, a segmentation map is generated to represent the probability of each pixel  belonging to text regions. However those text regions in score map always can not be separated from each other,  as a result of the overlapping of text regions and inaccurate predictions of text pixels. To get the text bounding boxes from the segmentation map, complex post-processing are conducted in \cite{zhang2016multi, yao2016scene}. 

Inspired by InstanceFCN \cite{dai2016instance}, we use position-sensitive segmentation to generate text segmentation maps. Compared with previous text segmentation methods, relative positions are generated. In detail, for a text bounding box $R$, a $g\times g$ regular grid is used to divide the text bounding  box into multiple bins (\emph{i.e.}, for a $2\times2$ grid, a text region can be split into $4$ bins, that is top-left, top-right, bottom-right, bottom-left). For each bin, a segmentation map is used to determine whether the pixels in this map belong to this bin.

We build position-sensitive segmentation with corner point detection in a unified network. We reuse the features of {$F_{3}$, $F_{4}$, $F_{7}$, $F_{8}$, $F_{9}$} and build some convolutional blocks  on them follow the residual block architecture of corner point detection branch (Shown in Fig. \ref{img_network}). All outputs of those blocks are resized to the scale of $F_{3}$ by bilinear upsampling with the scale factors set to {$1$, $2$, $4$, $8$, $16$}. Then all those outputs with the same scale are added together to generate richer features. We further enlarge the resolution of fused features by two continuous \emph{Conv1x1-BN-ReLU-Deconv2x2} blocks and set the kernels of the last deconvolution layer to $g\times g$. So, the final position-sensitive segmentation maps have $g \times g$ channels and the same size as the input images. In this work, we set $g$ to 2 in default.

\section{Training and Inference}

\subsection{Training}
\subsubsection{Label Generation}
\label{sec_label}

For an input training sample, we first convert each text box in ground truth into a rectangle that covers the text box region with minimal area and then determine the relative position of $4$ corner points.

We determine the relative position of a rotated rectangle by the following rules: 1) the x-coordinates of top-left and bottom-left corner points must less than the x-coordinates of top-right and bottom-right corner points; 2) the y-coordinates of top-left and top-right corner points must less than the y-coordinates of bottom-left and bottom-right corner points.  After that, the original ground truth can  be represented as a rotated rectangle with relative position of corner points. For convenience, we term the rotated rectangle $R=\{P_{i}|i\in\{1,2,3,4\}\}$, where $P_{i} = (x_{i}, y_{i})$ are the corner points of the  rotated rectangle  in top-left, top-right, bottom-right, bottom-left order.  

We generate the label of corner point detection and position-sensitive segmentation using  $R$. For corner point detection, we first compute the short side of $R$ and represent the $4$ corner points by  horizontal squares as shown in Fig. \ref{img_gt} (a).  For position-sensitive segmentation, we generate pixel-wise masks of text/non-text with $R$. We first initialize $4$ masks with the same scale as the input image and set all pixel value to $0$. Then we divide $R$ into four bins with a $2\times2$ regular grid and assign each bin to a mask, such as top-left bin to the first mask. After that, we set the value of  all pixels in those bins to $1$, as shown in Fig. \ref{img_gt} (b).

\begin{figure}

\begin{centering}
\includegraphics[scale=0.50]{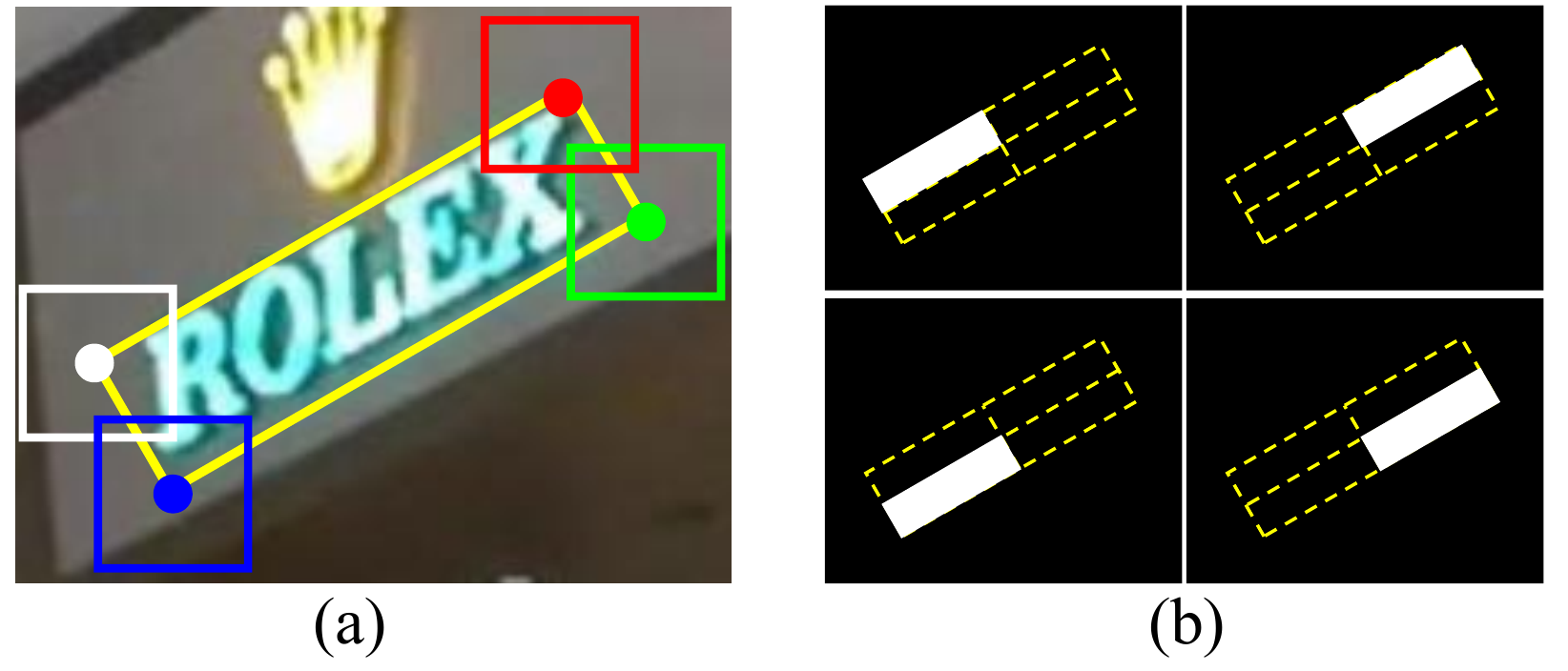}
\par\end{centering}
\caption{Label generation for corner points detection and position-sensitive segmentation. (a) Corner points are redefined and represented by  squares (boxes in white, red, green, blue) with the side length set as the short side of text bounding box $R$ (yellow box). (b) Corresponding ground truth of $R$ in (a) for position-sensitive segmentation.}
\label{img_gt}
\end{figure}

\subsubsection{Optimization}
We train the corner detection and position-sensitive segmentation simultaneously. The loss function is defined as:
\begin{equation}
L=\frac{1}{N_{c}}L_{conf}+\frac{\lambda_{1}}{N_{c}}L_{loc} + \frac{\lambda_{2}}{N_{s}}L_{seg}
\end{equation}
Where $L_{conf}$ and $L_{loc}$ are the loss functions of the score branch for predicting confidence score  and the offset branch for localization in the module of corner point detection. $L_{seg}$ is the loss function of position-sensitive segmentation. $N_{c}$ is the number of positive default boxes, $N_{s}$ is the number of pixels in segmentation maps.  $N_{c}$ and $N_{s}$ are used to normalize the losses of corner point detection and segmentation. $\lambda_{1}$ and $\lambda_{2}$ are the balancing factors of the three tasks. In default, we set the $\lambda_{1}$ to 1 and $\lambda_{2}$  to 10.

We follow the matching strategy of SSD and train the score branch using Cross Entropy loss:

\begin{equation}
L_{conf}=CrossEntropy(y_{c}, p_{c})
\end{equation}

Where $y_{c}$ is the ground truth of all default boxes, 1 for positive and 0 otherwise. $p_{c}$ is the predicted scores. In consideration of the extreme imbalance between positive and negative samples, the category homogenization is necessary. We use the online hard negative mining proposed in \cite{shrivastava2016training} to balance training samples and set the ratio of positives to negatives to $1:3$.

For the offset branch, we regress the offsets relative to  default boxes as Fast RCNN  \cite{Girshick_2015_ICCV} and optimize them with Smooth L1 loss:
\begin{equation}
L_{loc}=SmoothL1(y_{l},p_{l})
\end{equation}

Where $y_{l}=(\triangle x, \triangle y, \triangle s_{s}, \triangle s_{s})$ is the ground truth of offset branch and $p_{l}=(\triangle\tilde{x}, \triangle\tilde{y}, \triangle\tilde{s_{s}}, \triangle\tilde{s_{s}})$ is the predicted offsets. The $y_{l}$ can be calculated by a default box $B=(x_{b}, y_{b}, ss_{b}, ss_{b})$ and a corner point box $C=(x_{c}, y_{c}, ss_{c}, ss_{c})$:

\begin{equation}
\triangle x=\frac{x_{b}-x_{c}}{ss_{b}}  
\end{equation}

\begin{equation}
\triangle y=\frac{y_{b}-y_{c}}{ss_{b}}
\end{equation}

\begin{equation}
\triangle ss=\log(\frac{ss_{b}}{ss_{c}})
\end{equation}
 
We train position-sensitive segmentation by minimizing the Dice loss \cite{milletari2016v}:
\begin{equation}
L_{seg}=1-\frac{2y_{s}p_{s}}{y_{s} + p_{s}}
\end{equation}
Where $y_{s}$ is the label of position-sensitive segmentation and $p_{s}$ is the prediction of our segmentation module.

\subsection{Inference}
\subsubsection{Sampling and Grouping}
In inference stage, many corner points are yielded with the predicted location, short side and confidence score. Points with high score (great than 0.5 in default) are kept. After NMS, 4 corner point sets are composed  based on relative position information.

We generate the candidate bounding boxes by sampling and grouping the predicted corner points.  In theory, a rotated rectangle can be constructed by two points and a side perpendicular to the line segment made up by the two points. For a predicted point, the short side is known, so we can form a   
rotated rectangle by sampling and grouping two corner points in corner point sets arbitrarily, such as (top-left, top-right), (top-right, bottom-right), (bottom-left, bottom-right) and (top-left, bottom-left) pairs.

Several priori rules are used to filter unsuitable pairs: 1) the relative positional relations can not be violated, such as the x-coordinate of top-left point must less than that of top-right point in (top-left, top-right) pair; 2) the shortest side of the constructed rotated rectangle must be greater than a threshold (the default is 5); 3) the predicted short sides $ss_{1}$ and $ss_{2}$ of the two points in a pair must satisfy:
\begin{equation}
\frac{\max(ss_{1},ss_{2})}{\min(ss_{1},ss_{2})}\leq1.5
\end{equation}

\subsubsection{Scoring}
\label{sec_scoring}
A large number of candidate bounding boxes can be generated after sampling and 
grouping corner points. Inspired by InstanceFCN\cite{dai2016instance} and RFCN \cite{dai2016r}, we score the candidate bounding boxes by the position-sensitive segmentation maps. The  processes are shown in Fig. \ref{img_gt}.

To handle the rotated text bounding boxes, we adapt the Position-Sensitive ROI pooling layer in \cite{dai2016r} and propose \textbf{Rotated Position-Sensitive ROI Average pooling layer}. Specifically, for a rotated box, we first split the box into $g\times g$ bins. Then we generate a rectangle for each bin with the minimum area to cover the bin. We loop over all pixels in the minimum rectangle and calculate mean value of all pixels which in the bin. In the end, the score of a rotated bounding box is obtained by  averaging the means of $g\times g$ bins. The specific processes are shown in Algorithm \ref{algo_rps}. 

\begin{figure}
\begin{centering}
\includegraphics[scale=0.25]{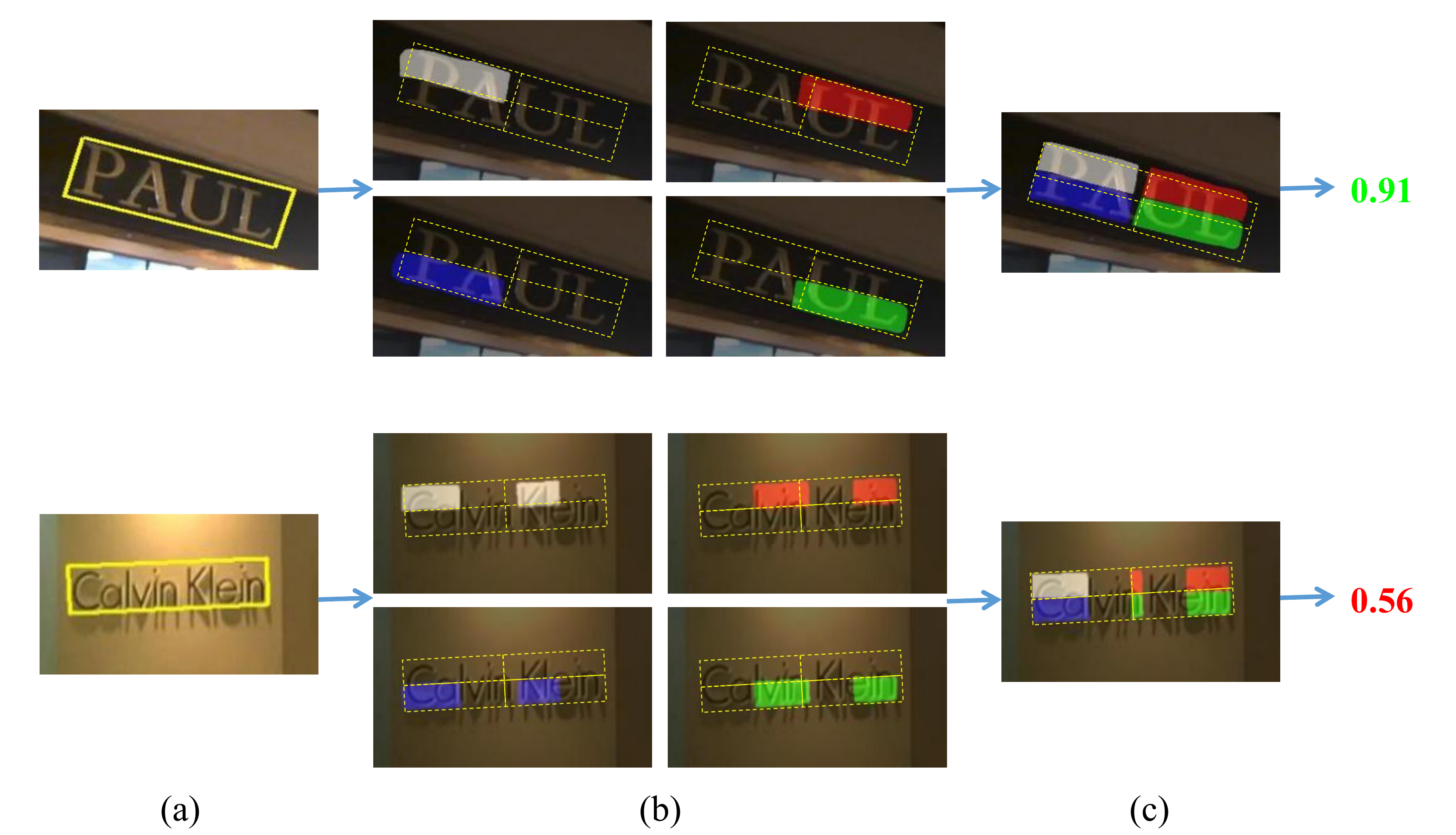}
\par\end{centering}
\caption{Overview of the scoring process. The yellow boxes in (a) are candidate boxes. (b) are predicted segmentation maps. We generate instance segment (c) of candidate boxes by assembling the segmentation maps as \cite{dai2016instance}. Scores are calculated by averaging the instance segment regions. }
\label{img_gt}
\vspace{-3mm}
\end{figure}

\begin{algorithm}[t]
\caption{Rotated Position-Sensitive ROI Average Pooling} 
\hspace*{0.02in} {\bf Input:} 
rotated bounding box $B$, $g\times g$ regular grid $G$, Segmentation maps $S$ 
\begin{algorithmic}[1]
\State Generating $Bins$ by spitting $B$ with $G$. 
\State $M\leftarrow 0$, $i\leftarrow 0$
\For{$i$ in $range(g\times g)$} 
	\State $bin\leftarrow Bins[i]$, $C\leftarrow 0$, $P\leftarrow 0$, 	\State $R \leftarrow MiniRect(bin)$ 
	\For{$pixel$ in $R$}
		\If{$pixel$ in $bin$} 
　　　　		\State $C\leftarrow C + 1$, $P\leftarrow P + G[i][pixel].value$
		\EndIf
	\EndFor
	\State $M\leftarrow M + \frac{P}{C}$
\EndFor
\State $score \leftarrow \frac{M}{g\times g}$
\State \Return score
\end{algorithmic}
\label{algo_rps}
\end{algorithm}

The candidate boxes with low score will be filtered out. We set the threshold $\tau$ to 0.6 by default.

\section{Experiments}
To validate the effectiveness of the proposed method, we conduct experiments on five public datasets: ICDAR2015, ICDAR2013, MSRA-TD500, MLT, COCO-Text, and compare with other state-of-the-art methods.

\subsection{Datasets}
\textbf{SynthText} \cite{gupta2016synthetic} is a synthetically generated dataset which consists of about 800000 synthetic images.   We use the dataset with word level labels to pre-train our model.

\textbf{ICDAR2015} is a dataset proposed in the Challenge 4 of the 2015 Robust Reading Competition \cite{karatzas2015icdar} for incidental scene text detection. There are 1000 images for training  and 500 images for testing with annotations labeled as word level quadrangles.

\textbf{ICDAR2013} is a dataset proposed in the Challenge 2  of the 2013 Robust Reading Competition \cite{karatzas2013icdar} focuses on horizontal text in scene. It contains 229 images for training and 233 images for testing.

\textbf{MSRA-TD500} \cite{yao2012detecting} is a dataset collected for detecting arbitrary-oriented long text lines. It consists of 300 training images and 200 test images with text line level annotations.

\textbf{MLT}  is a dataset that proposed on ICDAR2017 Competition \cite{MLT-Challenge} and  focuses on the multi-oriented, multi-script and multi-lingual aspects of scene text. It consists of 7200 training images, 2000 validation images and 9000 test images. 

\textbf{COCO-Text} \cite{veit2016coco} is a large scale scene text dataset which comes from the MS COCO dataset \cite{lin2014microsoft}. There are 63686 images are annotated and two versions of annotations and splits (V1.1 and V1.4) are released by the official. Previous methods are all evaluated on V1.1 and the new V1.4 are used on ICDAR2017 Competition \cite{COCO-Text-Challenge}. 
\subsection{Implementation Details}
\textbf{Training} Our model is pre-trained on SynthText then finetuned on other datasets (except COCO-Text). We use Adam \cite{kingma2014adam} to optimize our model with the learning rate fixed to $1e-4$. In pre-train stage, we train our model on SynthText for one epoch. During finetuning stage, the number of iterations are decided by the sizes of datasets.

\textbf{Data Augmentation} We use the same way of data augmentation as SSD. We randomly sample a patch from the input image  in the manner of SSD, then resize the sampled patch to $512\times512$. 

\textbf{Post Processing} NMS is the only  post processing step of our method. We set the threshold of NMS to $0.3$.

Our method is implemented in PyTorch \cite{PyTorch}. All the experiments are conducted on a regular workstation (CPU: Intel(R) Xeon(R) CPU E5-2650 v3 @ 2.30GHz; GPU:Titan Pascal; RAM: 64GB). We train our model with the batch size of 24 on $4$ GPUs in parallel and evaluate our model on 1 GPU with batch size set as $1$.

\begin{figure*}

\noindent \begin{centering}
\includegraphics[scale=0.28]{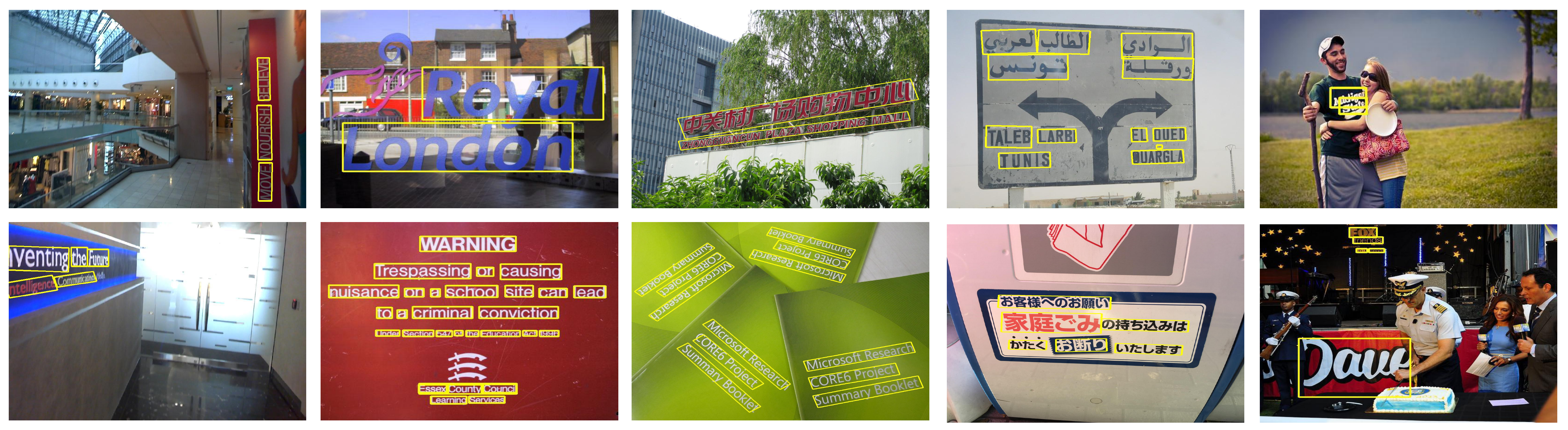}
\par\end{centering}
\caption{Examples of detection results. From left to right in columns: ICDAR2015, ICDAR2013, MSRA-TD500, MLT, COCO-Text.}
\label{img_results}
\end{figure*}

\subsection{Detecting Oriented Text}

We evaluate our model on the ICDAR2015 dataset to test its ability of arbitrarily oriented text detection. We finetune our model another 500 epochs on the datasets of ICDAR2015 and ICDAR2013. Note that, to detect vertical text better, in the last 15 epochs, we randomly rotate the sampled patches by $90$ degree or $-90$ degree with the probability  of $0.2$. In testing, we set $\tau$ to 0.7 and  resize the input images to $768\times1280$. Following \cite{Zhou_2017_CVPR,Hu_2017_ICCV,He_2017_ICCV}, we also evaluate our model on ICDAR2015 with multi-scale inputs, $\{512\times512, 768\times768, 768\times1280, 1280\times1280\}$ in default.

We compare our method with other state-of-the-art methods and list all the results in Table \ref{tab_icdar2015}. Our method outperforms the previous methods by a large margin. When tested at single scale, our method achieves the F-measure of $80.7\%$, which surpasses all competitors \cite{zhang2016multi,tian2016detecting,yao2016scene,Shi_2017_CVPR,Zhou_2017_CVPR,SSTD} . Our method achieves $84.3\%$ in F-measure with multi-scale inputs, higher than the current best one \cite{He_2017_ICCV} by $3.3\%$.


To explore the gain between our method which detects corner points and the method which regresses text boxes directly, we train a network named "baseline" in Table. \ref{tab_icdar2015} using the same settings as our method. The baseline model consists of the same backbone as our method and the prediction module in SSD/DSSD. With slight time cost, our method boost the accuracy greatly ($53.3\%$ \emph{vs} $80.7\%$).

\begin{table}
\small

\begin{centering}
\begin{tabular}{|c|c|c|c|c|}
\hline 
\textbf{Method} & \textbf{Precision} & \textbf{Recall} & \textbf{F-measure} & \textbf{FPS} \tabularnewline
\hline 
\hline 
Zhang \emph{et al.} \cite{zhang2016multi} & 70.8 & 43.0 & 53.6 & 0.48 \tabularnewline
\hline 
CTPN \cite{tian2016detecting} & 74.2  & 51.6  & 60.9 & 7.1 \tabularnewline
\hline 
Yao \emph{et al.} \cite{yao2016scene} & 72.3  & 58.7  & 64.8 & 1.61 \tabularnewline


\hline 
SegLink \cite{Shi_2017_CVPR} & 73.1  & \textbf{76.8} & 75.0 & - \tabularnewline
\hline 
EAST \cite{Zhou_2017_CVPR} & 80.5  & 72.8 & 76.4 & 6.52\tabularnewline
\hline 
SSTD \cite{SSTD} & 80.0  & 73.0 & 77.0 & \textbf{7.7} \tabularnewline

\hline
\textbf{baseline} & 66.0 & 44.7 & 53.3 & 4.5 \tabularnewline

\hline
\textbf{ours} & \textbf{94.1} & 70.7 & \textbf{80.7} & 3.6 \tabularnewline
\hline 
\hline 
EAST $^*$ $^\text{\dag}$ \cite{Zhou_2017_CVPR}& 83.3  & 78.3 & 80.7 & - \tabularnewline
\hline
WordSup $^*$ \cite{Hu_2017_ICCV} & 79.3  & 77.0  & 78.2 & \textbf{2} \tabularnewline
\hline 
He \emph{et al.} $^*$ $^\text{\dag}$ \cite{He_2017_ICCV} & 82.0  &  \textbf{80.0} & 81.0 & 1.1 \tabularnewline
\hline 
\textbf{ours}$^*$ & \textbf{89.5} & 79.7 & \textbf{84.3} & 1 \tabularnewline
\hline
\end{tabular}
\par\end{centering}
\caption{Results on ICDAR2015. $^*$ means multi-scale, $^\text{\dag}$ stands for the base net of the model is not VGG16.}
\label{tab_icdar2015}
\end{table}

\subsection{Detecting Horizontal Text}

We evaluate the ability of our model to detect horizontal text on ICDAR2013 dataset. We further train our model on ICDAR2013 dataset for 60 epochs on the basis of the finetuned ICDAR2015 model. In testing, the input images are resized to $512\times512$. We also use multi-scale inputs to evaluate our model.

The  results are listed in Table \ref{tab_icdar2013} and mostly are reported with the "Deteval" evaluation protocol. Our method achieves very competitive results. When tested at single scale, our method achieves the F-measure of $85.8\%$, which is slightly lower than the highest result. Besides, our method can run at 10.4 FPS, faster than most methods. For multi-scale evaluation, our method achieves the F-measure of $88.0\%$, which is also  competitive compared with other methods.

\begin{table}

\footnotesize
\begin{centering}
\begin{tabular}{|c|c|c|c|c|}
\hline 
\textbf{Method} & \textbf{Precision} & \textbf{Recall} & \textbf{F-measure} & \textbf{FPS} \tabularnewline
\hline 
\hline
Neumann \emph{et al.} \cite{neumann2015efficient} & 81.8 & 72.4 & 77.1 & 3  \tabularnewline

\hline 
Neumann \emph{et al.} \cite{neumann2016real} & 82.1 & 71.3  & 76.3  & 3 \tabularnewline

\hline
Fastext \cite{busta2015fastext} & 84.0 & 69.3 & 76.8 & 6  \tabularnewline
\hline 
Zhang \emph{et al.} \cite{zhang2015symmetry} & 88.0 & 74.0 & 80.0 & 0.02\tabularnewline
\hline
Zhang \emph{et al.} \cite{zhang2016multi} & 88.0  & 78.0  & 83.0 & 0.5 \tabularnewline

\hline 
Yao \emph{et al.} \cite{yao2016scene} & 88.9  & 80.2  & 84.3 & 1.61 \tabularnewline
\hline 
CTPN \cite{tian2016detecting} & 93.0  & 83.0  & \textbf{88.0} & 7.1 \tabularnewline

\hline
TextBoxes \cite{liao2017textboxes} & 88.0  & 74.0  & 81.0 & 11 \tabularnewline


\hline 
SegLink \cite{Shi_2017_CVPR} & 87.7  & 83.0 & 85.3 & \textbf{20.6}  \tabularnewline

\hline 
SSTD \cite{SSTD} & 89.0  & \textbf{86.0} & \textbf{88.0} & 7.7 \tabularnewline

\hline
\textbf{ours} & \textbf{93.3} & 79.4 & 85.8 & 10.4 \tabularnewline
\hline 
\hline
FCRN $^*$ \cite{gupta2016synthetic} & 92.0  & 75.5  & 83.0  & 0.8 \tabularnewline
\hline
TextBoxes $^*$ \cite{liao2017textboxes} & 89.0  & 83.0  & 86.0 & 1.3 \tabularnewline
\hline 
He \emph{et al.} $^*$ $^\text{\dag}$ \cite{He_2017_ICCV} & 92.0  &  81.0 & 86.0 & 1.1  \tabularnewline
\hline
WordSup $^*$ \cite{Hu_2017_ICCV} & \textbf{93.3}  & \textbf{87.5}  &  \textbf{90.3} & \textbf{2} \tabularnewline
\hline 
\textbf{ours}$^*$ & 92.0 & 84.4 & 88.0 & 1\tabularnewline
\hline
\end{tabular}
\par\end{centering}
\caption{Results on ICDAR2013. $^*$ means multi-scale, $^\text{\dag}$ stands for the base net of the model is not VGG16. Note that, the methods of the top three lines are evaluated under the "ICDAR2013" evaluation protocol.}
\label{tab_icdar2013}
\end{table}

\subsection{Detecting Long Oriented Text Line}

On MSRA-TD500, we evaluate the performance of our method for detecting long and multi-lingual text lines. HUST-TR400 is also used as training data as the MSRA-TD500 only contains 300 training images. The model is initialized with the model pre-trained on SynthText and then finetuned another 240 epochs. In test stage, we input the images with the size $768\times 768$ and set $\tau$ to 0.65.

As shown in Table \ref{tab_msra}, our method surpasses all the previous methods by a large margin. Our method achieves state-of-the-art performances both in recall, precision and F-measure ($87.6\%$, $76.2\%$ and $81.5\%$), and much better than the previous best result ($81.5\%$ \emph{vs.} $77.0\%$). That means our method is more capable than other methods of detecting arbitrarily oriented long text.

\begin{table}

\small

\begin{centering}
\begin{tabular}{|c|c|c|c|c|}
\hline 
\textbf{Method} & \textbf{Precision} & \textbf{Recall} & \textbf{F-measure} &\textbf{FPS} \tabularnewline
\hline 
\hline
TD-ICDAR \cite{yao2012detecting} & 53.0 & 52.0 & 50.0 & -\tabularnewline

\hline
TD-Mixture \cite{yao2012detecting} & 63.0 & 63.0 & 60.0 & -\tabularnewline

\hline
Kang \emph{et al.} \cite{kang2014orientation} & 71.0 &  62.0 & 66.0 & -\tabularnewline


\hline
Zhang \emph{et al.} \cite{zhang2016multi} & 83.0 & 67.0 & 74.0 & 0.48\tabularnewline

\hline 
Yao \emph{et al.} \cite{yao2016scene} & 76.5  & 75.3  & 75.9 &  1.61\tabularnewline


\hline 
EAST \cite{Zhou_2017_CVPR}  & 81.7  & 61.6 & 70.2 & 6.52 \tabularnewline
\hline 
EAST  $^\text{\dag}$ \cite{Zhou_2017_CVPR} & 87.3  & 67.4 & 76.1 & \textbf{13.2} \tabularnewline
\hline 
SegLink \cite{Shi_2017_CVPR} & 86.0  & 70.0 & 77.0 & 8.9 \tabularnewline
\hline 
He \emph{et al.} $^\text{\dag}$ \cite{He_2017_ICCV} & 77.0  &  70.0 & 74.0 & 1.1 \tabularnewline
\hline
\textbf{ours} & \textbf{87.6} & \textbf{76.2} & \textbf{81.5} & 5.7 \tabularnewline
\hline
\end{tabular}
\par\end{centering}
\caption{Results on MSRA-TD500. $^\text{\dag}$ stands for the base net of the model is not VGG16.}
\label{tab_msra}
\end{table}

\subsection{Detecting Multi-Lingual Text}

We verify the ability of our method to detect multi-lingual text on MLT. We finetune about 120 epochs on the model pre-trained on SynthText. When testing in single scale, the sizes of images are set as  $768 \times 768$. We evaluate our method online and compare with some public results on the leaderboard \cite{MLT-Challenge}. As shown in Table \ref{tab_mlt}, our method outperforms all competing methods by at least $3.1\%$. 

\begin{table}

\small

\begin{centering}
\begin{tabular}{|c|c|c|c|c|}
\hline 
\textbf{Method} & \textbf{Precision} & \textbf{Recall} & \textbf{F-measure}  \tabularnewline
\hline 
\hline
TH-DL \cite{MLT-Challenge} & 67.8 & 34.8 & 46.0 \tabularnewline
\hline
SARI\_FDU\_RRPN\_V1 \cite{MLT-Challenge} &71.2 & 55.5 & 62.4 \tabularnewline
\hline
Sensetime OCR \cite{MLT-Challenge} &56.9 & 69.4 & 62.6  \tabularnewline
\hline
SCUT\_DLVClab1 \cite{MLT-Challenge} &80.3  &54.5  &65.0  \tabularnewline

\hline
e2e\_ctc01\_multi\_scale \cite{MLT-Challenge} & 79.8 & 61.2 & 69.3 \tabularnewline

\hline
\textbf{ours} & \textbf{83.8} & 55.6 & 66.8  \tabularnewline
\hline
\textbf{ours}$^*$ & 74.3 & \textbf{70.6} & \textbf{72.4}  \tabularnewline
\hline
\end{tabular}
\par\end{centering}
\caption{Results on MLT. $^*$ means multi-scale.}
\label{tab_mlt}
\vspace{-3mm}
\end{table}

\subsection{Generalization Ability}

To evaluate the generalization ability of our model, we test it on COCO-Text using the model finetuned on ICDAR2015. We set the test image size as $768\times 768$. We use the annotations (V1.1) to compare with other methods, for the sake of fairness. The results are shown in Table \ref{tab_coco}. \textbf{Without training}, on COCO-Text, our method achieves an F-measure of $42.5\%$, better than competitors.

Besides, we also evaluate our model on the ICDAR2017 Robust Reading Challenge on COCO-Text \cite{COCO-Text-Challenge} with the annotations V1.4. The results are reported in Table \ref{tab_coco}. Among all the public results in leaderboard \cite{COCO-Text-Challenge}, our method ranks the first. Especially when the threshold of iou is set to 0.75, the result that our method exceeds others in a large margin shows it can detect text more accurately.

\begin{table}
\small

\begin{centering}
\begin{tabular}{|c|c|c|c|}
\hline 
\textbf{Method} & \textbf{Precision} & \textbf{Recall} & \textbf{F-measure}  \tabularnewline
\hline 
\hline
Baseline A  \cite{veit2016coco} & 83.8 & 23.3 &  36.5\tabularnewline

\hline 
Baseline B \cite{veit2016coco} & \textbf{89.7}  & 10.7  & 19.1 \tabularnewline

\hline 
Baseline C \cite{veit2016coco} & 18.6  & 4.7 & 7.5 \tabularnewline
\hline 
Yao \emph{et al.} \cite{yao2016scene} & 43.2  & 27.1 & 33.3 \tabularnewline
\hline 
EAST \cite{Zhou_2017_CVPR}   & 50.4  & \textbf{32.4} & 39.5  \tabularnewline
\hline 
WordSup \cite{Hu_2017_ICCV} & 45.2  & 30.9 & 36.8 \tabularnewline
\hline 
SSTD \cite{SSTD} & 46.0  &  31.0 & 37.0 \tabularnewline
\hline 
\textbf{ours}    &69.9  & 26.2 & 38.1    \tabularnewline
\hline 

\textbf{ours}$^*$ & 61.9 & \textbf{32.4} & \textbf{42.5}  \tabularnewline

\hline
\multicolumn{4}{|c|}{COCO-Text Challenge (IOU 0.5)}\tabularnewline
\hline
UM \cite{COCO-Text-Challenge} & 47.6   & \textbf{65.5}  & 55.1   \tabularnewline
\hline
TDN\_SJTU\_v2 \cite{COCO-Text-Challenge} & 62.4   & 54.3  & 58.1   \tabularnewline

\hline
Text\_Detection\_DL \cite{COCO-Text-Challenge} & 60.1 & 61.8  & 61.4   \tabularnewline

\hline
\textbf{ours} & \textbf{72.5} & 52.9 & 61.1  \tabularnewline

\hline
\textbf{ours}$^*$ & 62.9 & 62.2 & \textbf{62.6}  \tabularnewline

\hline 
\multicolumn{4}{|c|}{COCO-Text Challenge (IOU 0.75)}\tabularnewline
\hline
Text\_Detection\_DL \cite{COCO-Text-Challenge} & 25.2 & 25.5  & 25.4  \tabularnewline
\hline
UM \cite{COCO-Text-Challenge} & 22.7   & 31.2  & 26.3   \tabularnewline

\hline
TDN\_SJTU\_v2 \cite{COCO-Text-Challenge} & 31.8   & 27.7  &  29.6   \tabularnewline

\hline
\textbf{ours} & \textbf{40.0} & 30.0 & 34.6  \tabularnewline

\hline
\textbf{ours}$^*$ & 35.1 & \textbf{34.8} & \textbf{34.9}  \tabularnewline

\hline
\end{tabular}
\par\end{centering}
\caption{Results on COCO-Text.  $^*$ means multi-scale.}
\label{tab_coco}
\vspace{-3mm}
\end{table}

\vspace{-0.1cm}

\subsection{Limitations}
\begin{figure}

\begin{centering}
\includegraphics[scale=0.25]{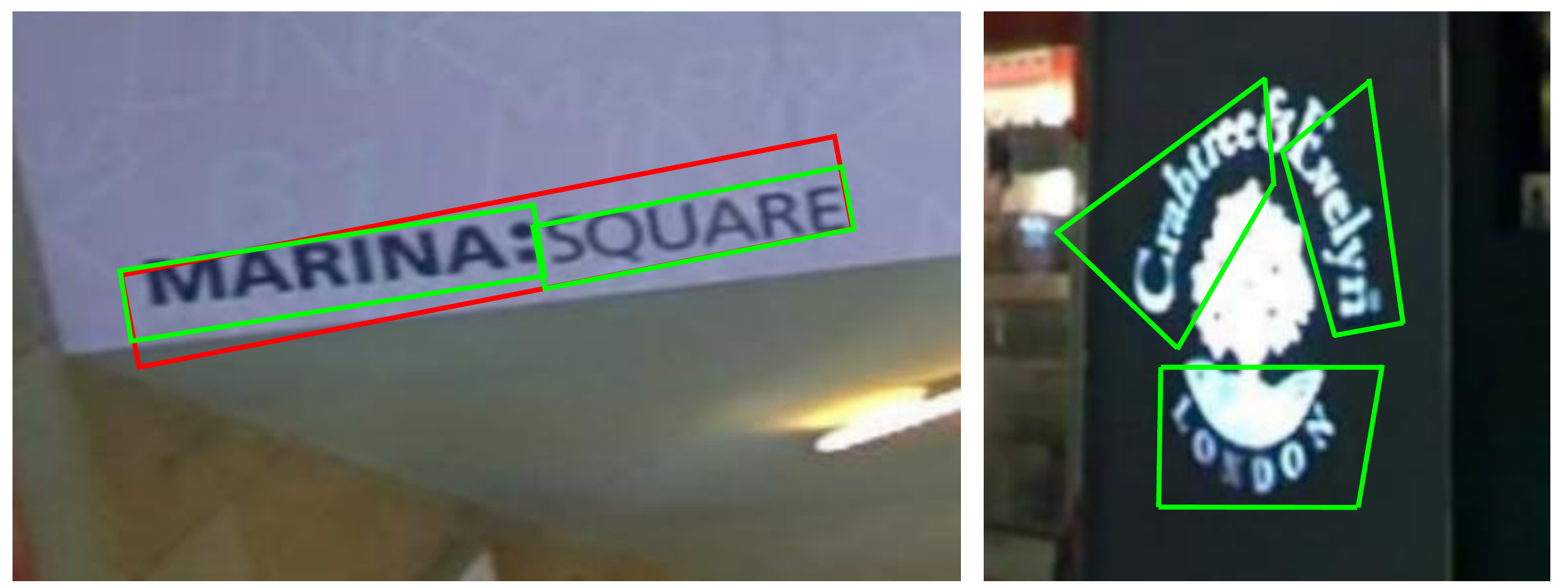}
\par\end{centering}
\caption{Failure cases of our method. The boxes in green are ground truth. The red boxes are our predictions.}
\label{img_limit}
\vspace{-4mm}
\end{figure}

\vspace{-0.1cm}
One limitation of the proposed method is that when two text instances are extremely close, it may predict the two instances as one (Fig. \ref{img_limit}), since the position-sensitive segmentation might fail. Besides, the method is not good at detecting curved text (Fig. \ref{img_limit}), as there are few curved samples in the training set.

\vspace{-0.1cm}
\section{Conclusion}
\vspace{-0.1cm}

In this paper, we have presented a scene text detector that localize text by corner point detection and position-sensitive segmentation. We evaluated it on several public benchmarks focusing on oriented, horizontal, long oriented and multi-lingual text. The superior performances demonstrate the effectiveness and robustness of our method. In the future, we are interested in constructing an end-to-end OCR system based on the proposed method.

\vspace{-1mm}
{
\small
\bibliographystyle{ieee}
\bibliography{egbib}

\begin{thebibliography}{10}\itemsep=-1pt

\bibitem{COCO-Text-Challenge}
Coco-text challenge.
\newblock \url{http://rrc.cvc.uab.es/?ch=5&com=evaluation&task=1}.

\bibitem{MLT-Challenge}
Mlt-challenge.
\newblock \url{http://rrc.cvc.uab.es/?ch=8&com=evaluation&task=1&gtv=1}.

\bibitem{PyTorch}
Pytorch.
\newblock \url{http://pytorch.org/}.

\bibitem{bai2017integrating}
X.~Bai, M.~Yang, P.~Lyu, Y.~Xu, and J.~Luo.
\newblock Integrating scene text and visual appearance for fine-grained image
  classification.
\newblock {\em arXiv preprint arXiv:1704.04613}, 2017.

\bibitem{bissacco2013photoocr}
A.~Bissacco, M.~Cummins, Y.~Netzer, and H.~Neven.
\newblock Photoocr: Reading text in uncontrolled conditions.
\newblock In {\em Proceedings of the IEEE International Conference on Computer
  Vision}, pages 785--792, 2013.

\bibitem{busta2015fastext}
M.~Busta, L.~Neumann, and J.~Matas.
\newblock Fastext: Efficient unconstrained scene text detector.
\newblock In {\em Proceedings of the IEEE International Conference on Computer
  Vision}, pages 1206--1214, 2015.

\bibitem{Busta_2017_ICCV}
M.~Busta, L.~Neumann, and J.~Matas.
\newblock Deep textspotter: An end-to-end trainable scene text localization and
  recognition framework.
\newblock In {\em Proc. ICCV}, 2017.

\bibitem{dai2016instance}
J.~Dai, K.~He, Y.~Li, S.~Ren, and J.~Sun.
\newblock Instance-sensitive fully convolutional networks.
\newblock In {\em European Conference on Computer Vision}, pages 534--549.
  Springer, 2016.

\bibitem{dai2016r}
J.~Dai, Y.~Li, K.~He, and J.~Sun.
\newblock R-fcn: Object detection via region-based fully convolutional
  networks.
\newblock In {\em Advances in neural information processing systems}, pages
  379--387, 2016.

\bibitem{epshtein2010detecting}
B.~Epshtein, E.~Ofek, and Y.~Wexler.
\newblock Detecting text in natural scenes with stroke width transform.
\newblock In {\em Computer Vision and Pattern Recognition (CVPR), 2010 IEEE
  Conference on}, pages 2963--2970. IEEE, 2010.

\bibitem{fu2017dssd}
C.-Y. Fu, W.~Liu, A.~Ranga, A.~Tyagi, and A.~C. Berg.
\newblock Dssd: Deconvolutional single shot detector.
\newblock {\em arXiv preprint arXiv:1701.06659}, 2017.

\bibitem{Girshick_2015_ICCV}
R.~Girshick.
\newblock Fast r-cnn.
\newblock In {\em The IEEE International Conference on Computer Vision (ICCV)},
  December 2015.

\bibitem{gomez2017textproposals}
L.~G{\'o}mez and D.~Karatzas.
\newblock Textproposals: a text-specific selective search algorithm for word
  spotting in the wild.
\newblock {\em Pattern Recognition}, 70:60--74, 2017.

\bibitem{gupta2016synthetic}
A.~Gupta, A.~Vedaldi, and A.~Zisserman.
\newblock Synthetic data for text localisation in natural images.
\newblock In {\em Proceedings of the IEEE Conference on Computer Vision and
  Pattern Recognition}, pages 2315--2324, 2016.

\bibitem{SSTD}
P.~He, W.~Huang, T.~He, Q.~Zhu, Y.~Qiao, and X.~Li.
\newblock Single shot text detector with regional attention.
\newblock In {\em The IEEE International Conference on Computer Vision (ICCV)},
  Oct 2017.

\bibitem{He_2017_ICCV}
W.~He, X.-Y. Zhang, F.~Yin, and C.-L. Liu.
\newblock Deep direct regression for multi-oriented scene text detection.
\newblock In {\em The IEEE International Conference on Computer Vision (ICCV)},
  Oct 2017.

\bibitem{Hu_2017_ICCV}
H.~Hu, C.~Zhang, Y.~Luo, Y.~Wang, J.~Han, and E.~Ding.
\newblock Wordsup: Exploiting word annotations for character based text
  detection.
\newblock In {\em The IEEE International Conference on Computer Vision (ICCV)},
  Oct 2017.

\bibitem{huang2015densebox}
L.~Huang, Y.~Yang, Y.~Deng, and Y.~Yu.
\newblock Densebox: Unifying landmark localization with end to end object
  detection.
\newblock {\em arXiv preprint arXiv:1509.04874}, 2015.

\bibitem{jaderberg2016reading}
M.~Jaderberg, K.~Simonyan, A.~Vedaldi, and A.~Zisserman.
\newblock Reading text in the wild with convolutional neural networks.
\newblock {\em International Journal of Computer Vision}, 116(1):1--20, 2016.

\bibitem{jaderberg2014deep}
M.~Jaderberg, A.~Vedaldi, and A.~Zisserman.
\newblock Deep features for text spotting.
\newblock In {\em European conference on computer vision}, pages 512--528.
  Springer, 2014.

\bibitem{kang2014orientation}
L.~Kang, Y.~Li, and D.~Doermann.
\newblock Orientation robust text line detection in natural images.
\newblock In {\em Proceedings of the IEEE Conference on Computer Vision and
  Pattern Recognition}, pages 4034--4041, 2014.

\bibitem{karatzas2015icdar}
D.~Karatzas, L.~Gomez-Bigorda, A.~Nicolaou, S.~Ghosh, A.~Bagdanov, M.~Iwamura,
  J.~Matas, L.~Neumann, V.~R. Chandrasekhar, S.~Lu, et~al.
\newblock Icdar 2015 competition on robust reading.
\newblock In {\em Document Analysis and Recognition (ICDAR), 2015 13th
  International Conference on}, pages 1156--1160. IEEE, 2015.

\bibitem{karatzas2013icdar}
D.~Karatzas, F.~Shafait, S.~Uchida, M.~Iwamura, L.~G. i~Bigorda, S.~R. Mestre,
  J.~Mas, D.~F. Mota, J.~A. Almazan, and L.~P. de~las Heras.
\newblock Icdar 2013 robust reading competition.
\newblock In {\em Document Analysis and Recognition (ICDAR), 2013 12th
  International Conference on}, pages 1484--1493. IEEE, 2013.

\bibitem{kingma2014adam}
D.~Kingma and J.~Ba.
\newblock Adam: A method for stochastic optimization.
\newblock {\em arXiv preprint arXiv:1412.6980}, 2014.

\bibitem{Li_2017_ICCV}
H.~Li, P.~Wang, and C.~Shen.
\newblock Towards end-to-end text spotting with convolutional recurrent neural
  networks.
\newblock In {\em The IEEE International Conference on Computer Vision (ICCV)},
  Oct 2017.

\bibitem{Li_2017_CVPR}
Y.~Li, H.~Qi, J.~Dai, X.~Ji, and Y.~Wei.
\newblock Fully convolutional instance-aware semantic segmentation.
\newblock In {\em The IEEE Conference on Computer Vision and Pattern
  Recognition (CVPR)}, July 2017.

\bibitem{liao2017textboxes}
M.~Liao, B.~Shi, X.~Bai, X.~Wang, and W.~Liu.
\newblock Textboxes: A fast text detector with a single deep neural network.
\newblock In {\em AAAI}, pages 4161--4167, 2017.

\bibitem{Lin_2017_CVPR}
T.-Y. Lin, P.~Dollar, R.~Girshick, K.~He, B.~Hariharan, and S.~Belongie.
\newblock Feature pyramid networks for object detection.
\newblock In {\em The IEEE Conference on Computer Vision and Pattern
  Recognition (CVPR)}, July 2017.

\bibitem{lin2014microsoft}
T.-Y. Lin, M.~Maire, S.~Belongie, J.~Hays, P.~Perona, D.~Ramanan,
  P.~Doll{\'a}r, and C.~L. Zitnick.
\newblock Microsoft coco: Common objects in context.
\newblock In {\em European conference on computer vision}, pages 740--755.
  Springer, 2014.

\bibitem{liu2016ssd}
W.~Liu, D.~Anguelov, D.~Erhan, C.~Szegedy, S.~Reed, C.-Y. Fu, and A.~C. Berg.
\newblock Ssd: Single shot multibox detector.
\newblock In {\em European conference on computer vision}, pages 21--37.
  Springer, 2016.

\bibitem{long2015fully}
J.~Long, E.~Shelhamer, and T.~Darrell.
\newblock Fully convolutional networks for semantic segmentation.
\newblock In {\em Proceedings of the IEEE Conference on Computer Vision and
  Pattern Recognition}, pages 3431--3440, 2015.

\bibitem{ma2017arbitrary}
J.~Ma, W.~Shao, H.~Ye, L.~Wang, H.~Wang, Y.~Zheng, and X.~Xue.
\newblock Arbitrary-oriented scene text detection via rotation proposals.
\newblock {\em arXiv preprint arXiv:1703.01086}, 2017.

\bibitem{milletari2016v}
F.~Milletari, N.~Navab, and S.-A. Ahmadi.
\newblock V-net: Fully convolutional neural networks for volumetric medical
  image segmentation.
\newblock In {\em 3D Vision (3DV), 2016 Fourth International Conference on},
  pages 565--571. IEEE, 2016.

\bibitem{neumann2010method}
L.~Neumann and J.~Matas.
\newblock A method for text localization and recognition in real-world images.
\newblock In {\em Asian Conference on Computer Vision}, pages 770--783.
  Springer, 2010.

\bibitem{neumann2015efficient}
L.~Neumann and J.~Matas.
\newblock Efficient scene text localization and recognition with local
  character refinement.
\newblock In {\em Document Analysis and Recognition (ICDAR), 2015 13th
  International Conference on}, pages 746--750. IEEE, 2015.

\bibitem{neumann2016real}
L.~Neumann and J.~Matas.
\newblock Real-time lexicon-free scene text localization and recognition.
\newblock {\em IEEE transactions on pattern analysis and machine intelligence},
  38(9):1872--1885, 2016.

\bibitem{redmon2016you}
J.~Redmon, S.~Divvala, R.~Girshick, and A.~Farhadi.
\newblock You only look once: Unified, real-time object detection.
\newblock In {\em Proceedings of the IEEE Conference on Computer Vision and
  Pattern Recognition}, pages 779--788, 2016.

\bibitem{ren2015faster}
S.~Ren, K.~He, R.~Girshick, and J.~Sun.
\newblock Faster r-cnn: Towards real-time object detection with region proposal
  networks.
\newblock In {\em Advances in neural information processing systems}, pages
  91--99, 2015.

\bibitem{Shi_2017_CVPR}
B.~Shi, X.~Bai, and S.~Belongie.
\newblock Detecting oriented text in natural images by linking segments.
\newblock In {\em The IEEE Conference on Computer Vision and Pattern
  Recognition (CVPR)}, July 2017.

\bibitem{shrivastava2016training}
A.~Shrivastava, A.~Gupta, and R.~Girshick.
\newblock Training region-based object detectors with online hard example
  mining.
\newblock In {\em Proceedings of the IEEE Conference on Computer Vision and
  Pattern Recognition}, pages 761--769, 2016.

\bibitem{simonyan2014very}
K.~Simonyan and A.~Zisserman.
\newblock Very deep convolutional networks for large-scale image recognition.
\newblock {\em arXiv preprint arXiv:1409.1556}, 2014.

\bibitem{tian2017wetext}
S.~Tian, S.~Lu, and C.~Li.
\newblock Wetext: Scene text detection under weak supervision.
\newblock {\em arXiv preprint arXiv:1710.04826}, 2017.

\bibitem{tian2015text}
S.~Tian, Y.~Pan, C.~Huang, S.~Lu, K.~Yu, and C.~Lim~Tan.
\newblock Text flow: A unified text detection system in natural scene images.
\newblock In {\em Proceedings of the IEEE international conference on computer
  vision}, pages 4651--4659, 2015.

\bibitem{tian2016detecting}
Z.~Tian, W.~Huang, T.~He, P.~He, and Y.~Qiao.
\newblock Detecting text in natural image with connectionist text proposal
  network.
\newblock In {\em European Conference on Computer Vision}, pages 56--72.
  Springer, 2016.

\bibitem{Tychsen-Smith_2017_ICCV}
L.~Tychsen-Smith and L.~Petersson.
\newblock Denet: Scalable real-time object detection with directed sparse
  sampling.
\newblock In {\em The IEEE International Conference on Computer Vision (ICCV)},
  Oct 2017.

\bibitem{veit2016coco}
A.~Veit, T.~Matera, L.~Neumann, J.~Matas, and S.~Belongie.
\newblock Coco-text: Dataset and benchmark for text detection and recognition
  in natural images.
\newblock {\em arXiv preprint arXiv:1601.07140}, 2016.

\bibitem{wang2011end}
K.~Wang, B.~Babenko, and S.~Belongie.
\newblock End-to-end scene text recognition.
\newblock In {\em Computer Vision (ICCV), 2011 IEEE International Conference
  on}, pages 1457--1464. IEEE, 2011.

\bibitem{wang2017point}
X.~Wang, K.~Chen, Z.~Huang, C.~Yao, and W.~Liu.
\newblock Point linking network for object detection.
\newblock {\em arXiv preprint arXiv:1706.03646}, 2017.

\bibitem{yao2012detecting}
C.~Yao, X.~Bai, W.~Liu, Y.~Ma, and Z.~Tu.
\newblock Detecting texts of arbitrary orientations in natural images.
\newblock In {\em Computer Vision and Pattern Recognition (CVPR), 2012 IEEE
  Conference on}, pages 1083--1090. IEEE, 2012.

\bibitem{yao2016scene}
C.~Yao, X.~Bai, N.~Sang, X.~Zhou, S.~Zhou, and Z.~Cao.
\newblock Scene text detection via holistic, multi-channel prediction.
\newblock {\em arXiv preprint arXiv:1606.09002}, 2016.

\bibitem{zhang2015symmetry}
Z.~Zhang, W.~Shen, C.~Yao, and X.~Bai.
\newblock Symmetry-based text line detection in natural scenes.
\newblock In {\em Proceedings of the IEEE Conference on Computer Vision and
  Pattern Recognition}, pages 2558--2567, 2015.

\bibitem{zhang2016multi}
Z.~Zhang, C.~Zhang, W.~Shen, C.~Yao, W.~Liu, and X.~Bai.
\newblock Multi-oriented text detection with fully convolutional networks.
\newblock In {\em Proceedings of the IEEE Conference on Computer Vision and
  Pattern Recognition}, pages 4159--4167, 2016.

\bibitem{Zhou_2017_CVPR}
X.~Zhou, C.~Yao, H.~Wen, Y.~Wang, S.~Zhou, W.~He, and J.~Liang.
\newblock East: An efficient and accurate scene text detector.
\newblock In {\em The IEEE Conference on Computer Vision and Pattern
  Recognition (CVPR)}, July 2017.

\end{thebibliography}
}

\end{document}